\documentclass[letterpaper]{article} 
\usepackage{aaai2026}  
\usepackage{times}  
\usepackage{helvet}  
\usepackage{courier}  
\usepackage[hyphens]{url}  
\usepackage{graphicx} 
\usepackage{natbib}  
\usepackage{caption} 
\usepackage{algorithm}
\usepackage{algorithmic}
\usepackage{booktabs}
\usepackage{amsmath}
\usepackage{tabularx}
\usepackage{makecell}
\usepackage{adjustbox}

\usepackage{newfloat}
\usepackage{listings}
\usepackage{multirow}
\usepackage{array}
\usepackage[utf8]{inputenc} 
\DeclareCaptionStyle{ruled}{labelfont=normalfont,labelsep=colon,strut=off} 
\floatstyle{ruled}
\newfloat{listing}{tb}{lst}{}
\floatname{listing}{Listing}
\nocopyright
\title{ViDia2Std: A Parallel Corpus and Methods for Low-Resource Vietnamese Dialect-to-Standard Translation}
\author {
    Khoa Anh Ta\textsuperscript{\rm 1, 2},
    Nguyen Van Dinh\textsuperscript{\rm 1, 2},
    Kiet Van Nguyen\textsuperscript{\rm 1, 2}\thanks{Corresponding author.}
}
\affiliations {
    \textsuperscript{\rm 1}Faculty of Information Science and Engineering, University of Information Technology, \\Ho Chi Minh City, Vietnam\\
    \textsuperscript{\rm 2}Vietnam National University, Ho Chi Minh City, Vietnam\\
    \{21522232, 20520657\}@gm.uit.edu.vn, kietnv@uit.edu.vn
}

\begin{document}

\maketitle

\begin{abstract}
Vietnamese exhibits extensive dialectal variation, posing challenges for NLP systems trained predominantly on standard Vietnamese. Such systems often underperform on dialectal inputs, especially from underrepresented Central and Southern regions. Previous work on dialect normalization has focused narrowly on Central-to-Northern dialect transfer using synthetic data and limited dialectal diversity. These efforts exclude Southern varieties and intra-regional variants within the North. We introduce ViDia2Std, the first manually annotated parallel corpus for dialect-to-standard Vietnamese translation covering all 63 provinces. Unlike prior datasets, ViDia2Std includes diverse dialects from Central, Southern, and non-standard Northern regions often absent from existing resources, making it the most dialectally inclusive corpus to date. The dataset consists of over 13,000 sentence pairs sourced from real-world Facebook comments and annotated by native speakers across all three dialect regions. To assess annotation consistency, we define a semantic mapping agreement metric that accounts for synonymous standard mappings across annotators. Based on this criterion, we report agreement rates of 86\% (North), 82\% (Central), and 85\% (South). We benchmark several sequence-to-sequence models on ViDia2Std. mBART-large-50 achieves the best results (BLEU 0.8166, ROUGE-L 0.9384, METEOR 0.8925), while ViT5-base offers competitive performance with fewer parameters. ViDia2Std demonstrates that dialect normalization substantially improves downstream tasks, highlighting the need for dialect-aware resources in building robust Vietnamese NLP systems.
\end{abstract}

\begin{links}
    \link{Code}{https://github.com/biuinvincible/ViDia2Std.git}
    \link{Datasets}{https://huggingface.co/datasets/Biu3010/ViDia2Std}
\end{links}

\section{Introduction}
\label{sec:introduction}

Vietnamese dialects pose significant challenges for NLP systems due to the mismatch between regionally diverse inputs and the standardized data on which most models are trained. This problem is especially acute in low-resource settings, where dialectal inputs are underrepresented and lack high-quality annotated corpora. 

Although we refer to the task as \textit{dialect-to-standard translation} in the title of this paper—to remain consistent with prior literature and emphasize the parallel nature of our dataset—we adopt the term \textit{dialect normalization} throughout the paper. This reflects our interpretation of the task as a preprocessing step that transforms non-standard regional language into standardized Vietnamese, enabling downstream models to process dialectal input more effectively. In the Vietnamese context, where dialects are variations within a single language rather than mutually unintelligible systems, normalization better captures the linguistic nature of this transformation.

As \citet{alam-anastasopoulos-2025-large} observe, ``NLP models trained on standardized language data often struggle with variations''. This is particularly true for Vietnamese, which comprises three major dialect groups—Northern, Central, and Southern—that differ substantially in phonology, vocabulary, and syntax \cite{le-luu-2023-parallel}. Standard Vietnamese is based on the Northern dialect, and most existing Vietnamese NLP models (e.g., PhoBERT, BARTpho) are trained predominantly on this variety. As a result, these models often fail to interpret utterances containing region-specific vocabulary and expressions.

Table~\ref{tab:example} illustrates the performance of both commercial translation systems and state-of-the-art language models (e.g., Claude Sonnet 4, Gemini-2.5-Flash, Google Translate, Bing Translator) when handling dialectal Vietnamese inputs. These models exhibit significant errors or misunderstandings, highlighting the urgent need for effective dialect normalization as a preprocessing step to improve translation and understanding.

Several prior efforts have attempted to address this problem. Notably, \citet{le-luu-2023-parallel} constructed a Central-to-Standard Vietnamese parallel corpus. However, their data was collected exclusively from speakers in Ha Tinh province, thus limiting the geographical and lexical diversity represented. Central dialects vary widely—provinces like Thua Thien Hue, Quang Tri, and Thanh Hoa each contribute distinct vocabulary and syntactic patterns. Moreover, regional variation is not confined to the Central region: many provinces in both the North and South exhibit distinct expressions that challenge NLP tools trained solely on the Northern standard.

To overcome these limitations, we introduce \textbf{ViDia2Std}, the first large-scale Vietnamese dialect normalization corpus with nationwide coverage. ViDia2Std consists of over 13,000 manually aligned sentence pairs drawn from real user comments collected from all 63 provinces of Vietnam, ensuring diverse representation across Northern, Central, and Southern dialects. A rigorous annotation protocol involving native speakers from each dialect region ensures high-quality standardization.

\begin{table}[t] 
    \centering
    \includegraphics[width=1\linewidth]{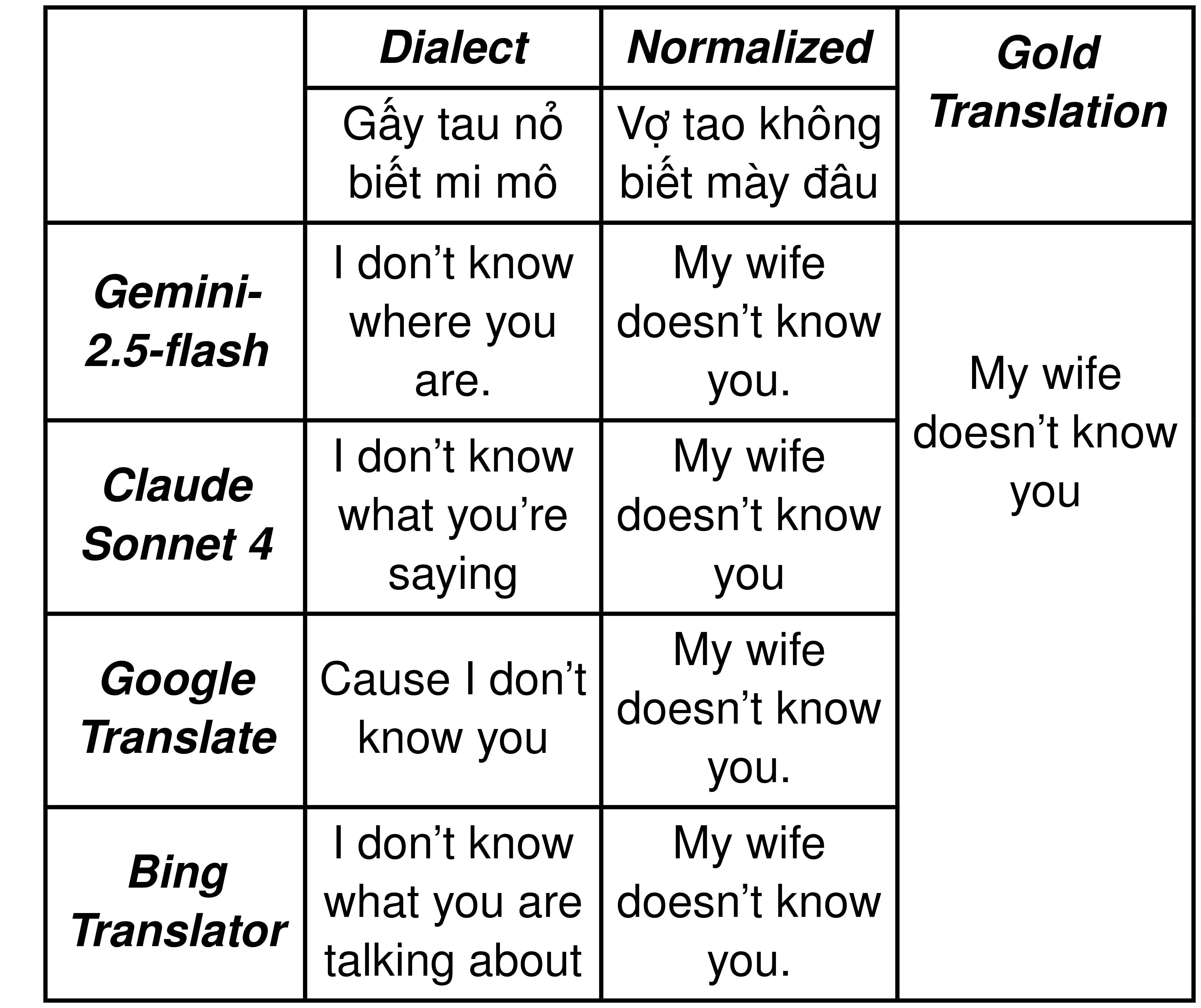}
    \caption{Comparison of Translation Models on Dialectal and Normalized Data.} 
    \label{tab:example}
\end{table}

In addition to releasing this corpus, we establish strong baselines using modern sequence-to-sequence models such as BARTpho and ViT5, fine-tuned for the dialect normalization task. These models demonstrate effective handling of dialectal variation and offer reference results for future research.

Importantly, we show that dialect normalization leads to substantial improvements on downstream tasks. For example, sentiment classification accuracy on dialectal inputs increases from 51\% to 62\% after normalization. Similarly, normalized inputs improve the performance of machine translation systems and large language models. These findings confirm that dialect normalization is a crucial preprocessing step for robust Vietnamese NLP.

\paragraph{Our contributions are as follows:}
\begin{itemize}
  \item \textbf{ViDia2Std corpus:} We release a high-quality, manually annotated Vietnamese dialect-to-standard parallel corpus covering all major dialect regions. The data—sourced from public Facebook comments nationwide—far exceeds prior resources in both size and dialectal diversity.
  \item \textbf{Baseline normalization models:} We evaluate several neural sequence-to-sequence architectures (e.g., BARTpho, ViT5) on the dialect normalization task, providing reference benchmarks for future work.
  \item \textbf{Downstream evaluation:} We demonstrate that dialect normalization improves core NLP tasks. For instance, our normalization models raise sentiment-analysis F1 (Weighted) from 0.52 to 0.63 and significantly enhance translation quality, confirming the benefit of integrating normalization into Vietnamese NLP pipelines.
\end{itemize}

\section{Related Work}
\label{sec:related_work}

Text normalization has been extensively studied in historical text restoration \cite{bollmann-2019-large}, progressing from rule-based systems to neural seq2seq models. In Vietnamese, ViLexNorm \cite{nguyen-etal-2024-vilexnorm} introduced a 10K-sentence corpus for standardizing informal online text, improving downstream tasks like POS tagging. However, dialect normalization is more complex, involving systematic variation in vocabulary, syntax, and phonology. Globally, recent work treats dialect normalization as character- or byte-level transduction \cite{kuparinen-etal-2023-dialect}, or multi-dialect machine translation \cite{abe-etal-2018-multi}, with growing use of adapters for robustness in low-resource dialects \cite{held-etal-2023-tada}. Ibn Alam et al.~\cite{alam-anastasopoulos-2025-large} show that fine-tuning open LLMs with small dialect corpora can significantly boost BLEU scores. In Vietnamese, \citet{le-luu-2023-parallel} proposed a parallel corpus for Central-to-North normalization and showed BARTpho outperforms multilingual models, but their dataset is limited in size and regional diversity.

\paragraph{Our Contribution.}
We introduce \textbf{ViDia2Std}, the first nationwide, manually annotated Vietnamese dialect-to-standard corpus covering all 63 provinces and three major dialect zones. Our work significantly scales previous efforts and evaluates multiple state-of-the-art normalization models. Beyond intrinsic evaluation, we conduct extrinsic tests on sentiment analysis and translation, confirming that normalization leads to large performance gains in real-world Vietnamese NLP applications

\section{Dataset Construction}
\label{sec:dataset_construction}
\begin{figure*}[t]
    \centering
    \includegraphics[width=1\textwidth]{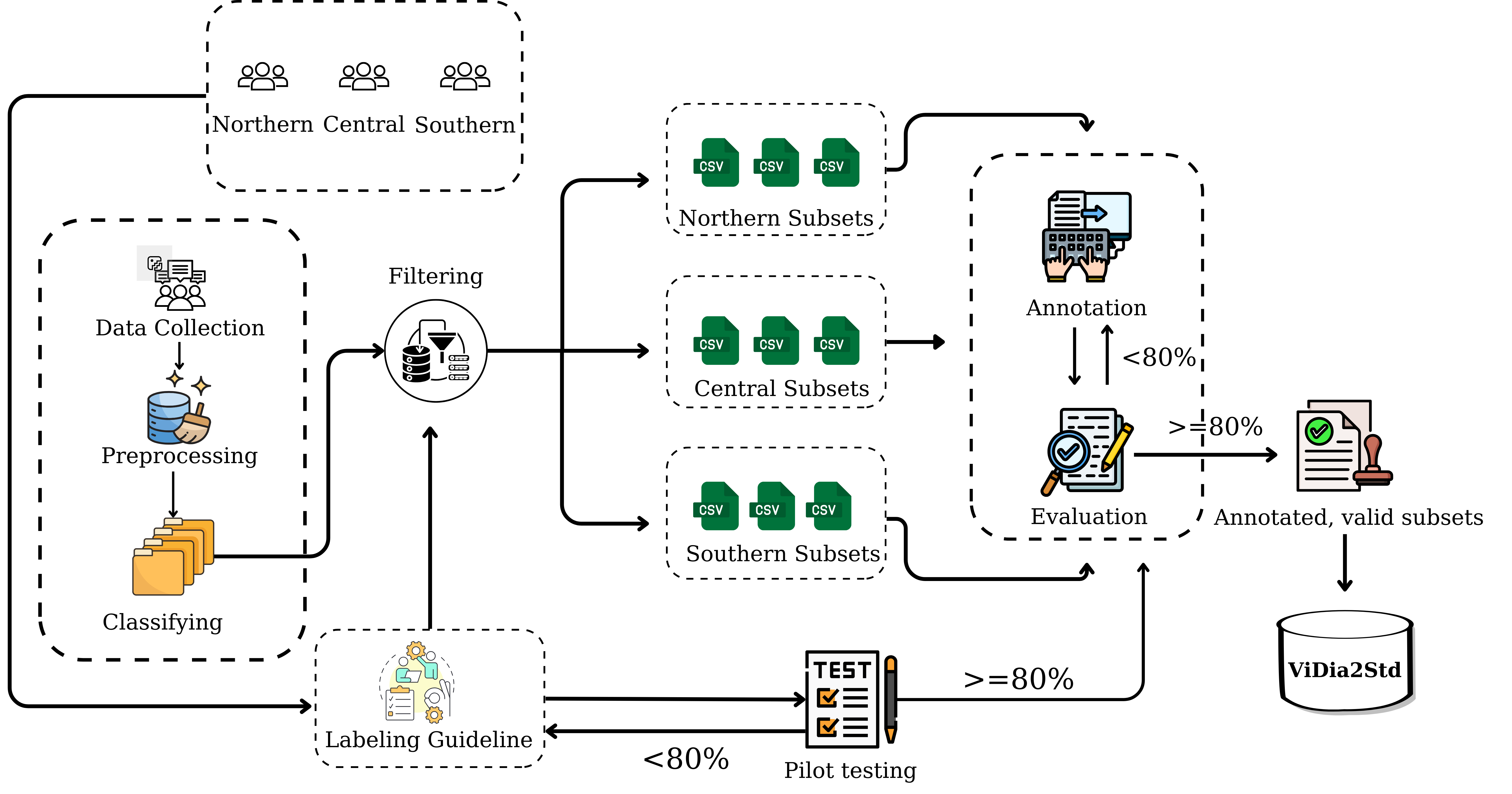}
    \caption{Overview of our dataset creation pipeline, from raw social media data to the final dialect-standard parallel corpus.}
    \label{fig:dataset_pipeline}
\end{figure*}

In this section, we describe the process of constructing our Vietnamese dialect-to-standard parallel corpus. The overall data creation pipeline is illustrated in Figure~\ref{fig:dataset_pipeline}.

\subsection{Data Collection}
\label{subsec:data_collection}
We construct a parallel corpus (dialect–standard) by harvesting user-generated content from Facebook. Our collection strategy involved identifying potential news fanpages for all 63 provinces of Vietnam. \textbf{We prioritized fanpages managed by individuals over those managed by local authorities, as the former exhibited a significantly higher rate of user interaction using local dialects}. Using Python scripts with the Selenium library, we then collected data. The target of 350-500 posts per province was a guideline, not a rigid quota, as fanpage activity varied greatly. \textbf{If a province yielded too few comments, we actively searched for other relevant local groups and fanpages to ensure fair representation and avoid bias toward any single province}.

Facebook is ideal because it is widely used across Vietnam (over 70\% of the population on social media in 2022 \cite{nguyen-etal-2024-vilexnorm}) and hosts open groups where users naturally write in local dialects. In line with prior work in other languages, we follow a “web-as-corpus” approach to dialectology \cite{burghardt-etal-2016-creating, alshutayri2019social}. For example, Alshutayri and Atwell built a multi-dialect Arabic corpus from Facebook and Twitter \cite{alshutayri2019social}, and Burghardt et al.\ used Bavarian Facebook posts to compile a dialect lexicon \cite{burghardt-etal-2016-creating}. Inspired by \cite{le-luu-2023-parallel}, who prepared scripted dialogues to elicit Central/Northern Vietnamese variants, we instead use authentic Facebook comments as prompts. Native speakers take each comment and restate it in an accurate, cleaned dialectal form (see Annotation below), then map it to standard Vietnamese. In this way we leverage real multilingual user data without incurring the cost of eliciting responses for all provinces.

\subsection{Data Preprocessing and Dialect Filtering}
\label{subsec:data_proprocess_filtering}
The raw social media data was highly unstructured and noisy, containing emojis, URLs, non-standard text, and metadata. To extract clean dialectal content suitable for annotation, we applied a \textbf{two-stage pipeline}:

\textbf{Stage 1: Automatic Preliminary Preprocessing.}
We first applied an automatic script to de-noise the raw comments. This involved several key steps:
\begin{itemize}
    \item Converting all text to lowercase.
    \item Removing non-linguistic content (e.g., empty comments, emojis, stickers, repeated characters).
    \item Stripping metadata (e.g., URLs, @mentions, \#hashtags).
    \item Normalizing common social media language and \textit{teencode} using a manually curated dictionary (e.g., "ko" --$>$ "khong", "ae" --$>$ "anh em").
\end{itemize}
This stage was designed to reduce noise and ease the subsequent human annotation burden.

\textbf{Stage 2: Annotator-Driven Dialect Filtering.}
The cleaned data from Stage 1 was then presented to our annotator team for dialect filtering. This step was crucial for isolating dialect-rich sentences from standard Vietnamese. Annotators used a set of \textbf{region-specific keyword lists}, prepared by our team based on linguistic analysis, to identify and extract comments with a high probability of containing dialect.
This two-stage pipeline yielded a cleaner, dialect-rich subset suitable for the main annotation task (described in the next subsection). 

\subsection{Annotation Protocol}
\label{subsec:annotation_protocol}

We recruited nine native Vietnamese annotators, purposefully selected to represent the three major dialect regions: North (2), Central (4), and South (3). The Central region was assigned more annotators due to its higher internal lexical diversity. All annotators received basic linguistic training and underwent cross-dialectal familiarization via dialect blogs, glossaries, and curated social media content.

Dialect normalization is treated as a lexical-semantic mapping task rather than free-form paraphrase. Each sentence is annotated in three stages:

\begin{enumerate}
    \item \textbf{Dialect Cleaning:} Annotators first fix orthographic issues (e.g., typos, abbreviations, missing punctuation) while retaining all dialectal lexical and syntactic features.
    \item \textbf{Dialect-to-Standard Mapping:} Dialectal tokens are translated into natural standard Vietnamese, preserving meaning, tone, and sentence structure.
    \item \textbf{Ambiguity Flagging:} Any cases that are idiomatic, ambiguous, or difficult to align are flagged for collaborative team discussion.
\end{enumerate}

To assist with rare or ambiguous cases, annotators are provided with dialect dictionaries, regional glossaries, and prior annotated corpora. This hybrid approach—combining native intuition with structured resources—helps ensure semantic fidelity across diverse dialectal input. Annotators were compensated at a rate of \$0.038 (1,000 VND) per sentence pair, broken down as \$0.019 for the manual normalization stage and \$0.019 for the dialect-to-standard labeling stage.

\subsection{Annotator Training and Quality Control}
\label{subsec:annotator_training_qc}

To maximize regional coverage, annotators from each dialect zone were exposed to lexical variation across neighboring provinces. For instance, Central annotators from Nghe An were introduced to features from Quang Tri and Hue to enable broader dialectal understanding. This intra-regional calibration improves coverage of localized vocabulary (e.g., ``bui'' for ``vui'' in Quang Tri) and enhances consistency within each macro-region.

Prior to full-scale annotation, we conducted a pilot round in which all annotators collectively annotated a shared set of 100 dialectal sentences. No single annotator or gold reference was used for comparison. Instead, we adopt a strict group-level semantic consistency criterion:

\textbf{A sentence is marked as “agreed” if and only if all annotators produce semantically equivalent mappings for every dialectal token.} Semantic equivalence is defined using curated synonym sets and reviewed during collaborative sessions.

Formally, let \( N \) be the total number of pilot sentences, and \( K \) be the number of annotators. For sentence \( i \), each dialectal token \( b_j \) is mapped to a normalized form \( c_j^{(k)} \) by annotator \( k \). The sentence is considered semantically agreed upon if:

\begin{equation}
\text{Agree}(i) = 
\begin{cases}
1, & \text{if } \forall j,~ \forall k_1, k_2,~ c_j^{(k_1)} \sim c_j^{(k_2)} \\
0, & \text{otherwise}
\end{cases}
\end{equation}

The overall \textbf{Strict Semantic Group Agreement (SSGA)} score is then computed as:

\begin{equation}
\text{SSGA} = \frac{1}{N} \sum_{i=1}^{N} \text{Agree}(i)
\end{equation}

Annotators were only allowed to proceed with full annotation if the group achieved an SSGA of at least 80\% on the pilot set. If this threshold was not met, \emph{all annotators were retrained collectively}, with clarification of disagreements and refinement of dialectal interpretation rules.

In the final corpus, group-level agreement scores reached 86\% (North), 82\% (Central), and 85\% (South), indicating high internal consistency. The strict nature of SSGA makes it more demanding than pairwise or majority-vote metrics but better reflects full-group semantic convergence.

\subsection{Corpus Summary and Release}
\label{subsec:corpus_sum_release}

The resulting corpus comprises 13,657 dialect–standard sentence pairs, each aligned at the sentence level. One column contains the cleaned dialectal sentence, and the other its semantically equivalent form in standard Vietnamese. Examples of these sentence pairs are illustrated in Table~\ref{tab:data_samples}.

The dataset will be released for non-commercial research purposes. We hope this resource will support future work in dialect-aware modeling, low-resource machine translation, and inclusive Vietnamese NLP.

\begin{table}[htbp] 
\centering
\includegraphics[width=1\columnwidth]{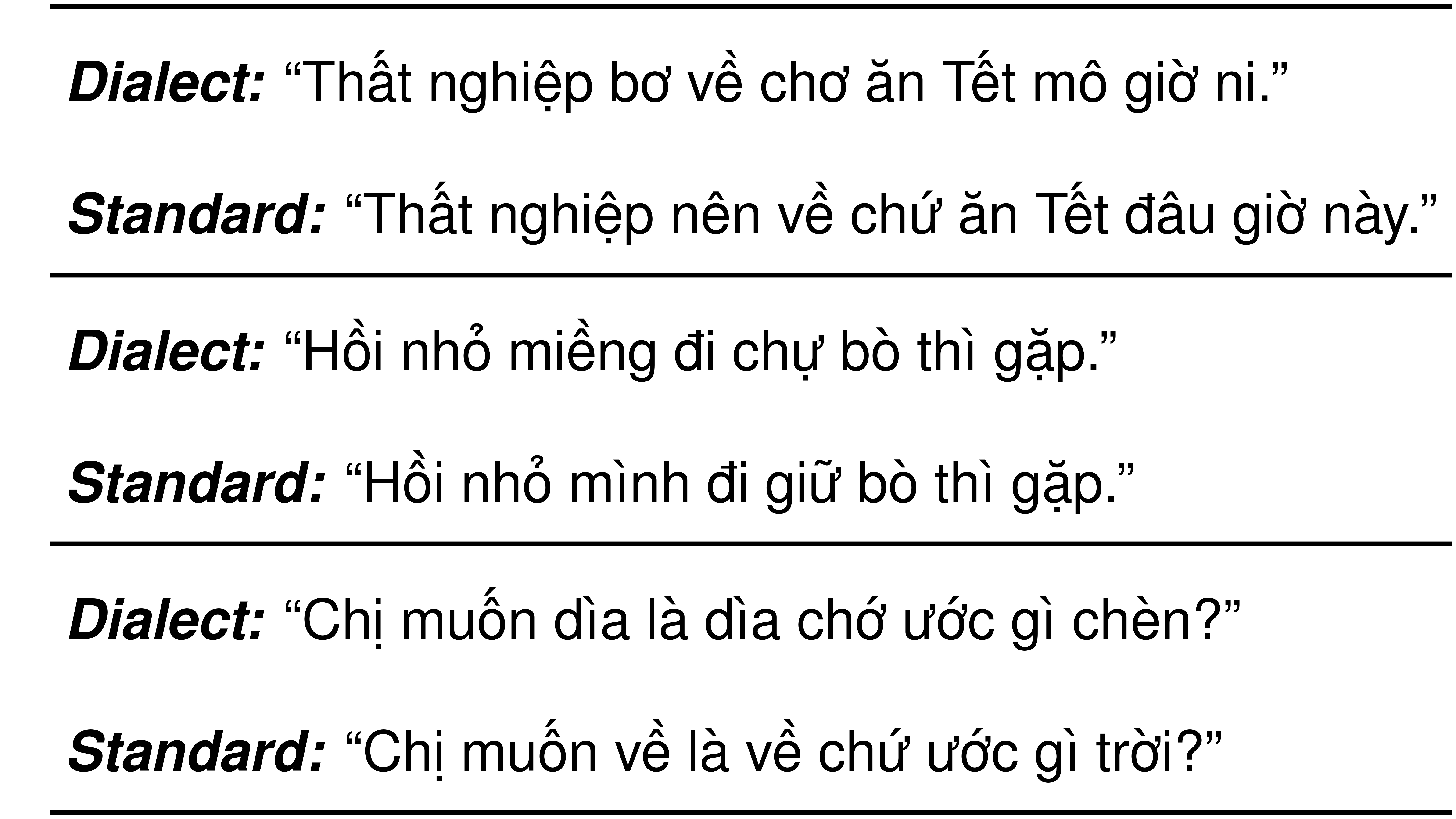}
\caption{Sample sentence pairs: dialect vs.\ standard} 
\label{tab:data_samples} 
\end{table}

\begin{table}[htbp]
\centering
\begin{tabular}{l c cc}
\toprule
\textbf{Region} & \textbf{Number of Sentences} & \multicolumn{2}{c}{\textbf{Avg. Words}} \\
\cmidrule(lr){3-4}
& & \textbf{Dialect} & \textbf{Standard} \\
\midrule
Central  & 9,033  & 10.55 & 10.82 \\
Northern & 1,054  & 9.31  & 9.28  \\
Southern & 3,570  & 10.58 & 11.02 \\
\midrule
\textbf{Total}   & \textbf{13,657} & \textbf{10.46} & \textbf{10.77} \\
\bottomrule
\end{tabular}
\caption{Basic Dataset Statistics}
\label{tab:basic_stats}
\end{table}

\section{Intrinsic Evaluation}
\label{sec:intrinsic_eval}

\subsection{Task Definition}
\label{subsec:task_definition}
Following the approach of \citet{le-luu-2023-parallel}, we define Vietnamese dialect normalization as a conditional sequence generation task. Given an input sentence $x = [x_1, x_2, \dots, x_n]$ in dialectal Vietnamese, the model generates an equivalent sentence $y = [y_1, y_2, \dots, y_m]$ in standard Vietnamese by optimizing the conditional probability $P(y|x)$:

\begin{equation}
L = -\sum_{t=1}^{m} \log P(y_t \mid y_{<t}, x; \theta) 
\end{equation}

where $\theta$ denotes model parameters. The objective is to preserve both meaning and pragmatic intent (e.g., interrogatives, imperatives), rather than perform simple word-level substitution.

\subsection{Experimental Setup}
\label{subsec:experimental_setup}
This subsection outlines the foundational elements of the experimental methodology, including the selection of models, the metrics employed for evaluation, and the specific training configurations.

\subsubsection{Models for Dialect Normalization}
\label{subsec:models_for_dialect_normalization}

We benchmark five sequence-to-sequence models fine-tuned for Vietnamese text generation:

\begin{itemize}
    \item \textbf{BARTpho-word-base} \cite{Tran2022BARTpho}: A Vietnamese BART model with word-level tokenization, capturing high-level grammatical structures.
    \item \textbf{BARTpho-syllable-base} \cite{Tran2022BARTpho}: A variant using syllable-level tokenization, better suited for Vietnamese’s monosyllabic nature and dialectal variations.
    \item \textbf{ViT5-base} \cite{phan-etal-2022-vit5}: A Vietnamese version of T5, treating all tasks as text-to-text generation; pre-trained on diverse NLP tasks for strong generalization.
    \item \textbf{Vietnamese-correction-v2} \cite{VietCorrection2024}: A BARTpho-syllable model trained for spelling correction; included to explore overlap with normalization tasks. \item \textbf{mBART-large-50} \cite{liu-etal-2020-multilingual-denoising,tang-etal-2020-multilingual}: A multilingual model trained on 50 languages via denoising objectives, offering strong robustness to diverse input forms including dialects.
\end{itemize}

\subsubsection{Evaluation Metrics}
\label{subsec:intrinsic_evaluation_metrics}

We evaluate model outputs using BLEU \cite{papineni-etal-2002-bleu}, ROUGE-L \cite{lin-2004-rouge}, METEOR \cite{banerjee-lavie-2005-meteor}, WER, and CER—capturing both surface-form accuracy and semantic preservation. This diverse metric set ensures reliable assessment for the nuanced task of dialect normalization.

\subsubsection{Training Configuration}
\label{subsec:intrinsic_training_configuration}

All models were trained using HuggingFace Transformers \cite{wolf-etal-2020-transformers}. The ViDia2Std corpus was split into 10870 samples for training, 1184 for development, and 1603 for testing.

To ensure a rigorous and fair comparison, \textbf{we applied an identical hyperparameter configuration across all models}. We used \texttt{AutoTokenizer} with a maximum sequence length of 50 tokens—sufficient given the average sentence length ($<$11 tokens). Training utilized the AdamW optimizer with a batch size of 32, a learning rate of 2e-5, and a maximum of 10 epochs, using mixed precision (\texttt{fp16=True}) on a single A100 GPU.

Models were evaluated and checkpointed each epoch, with early stopping (\texttt{patience=1}) based on BLEU improvements $\ge$0.01. Crucially, to ensure strict reproducibility, \textbf{all models in this camera-ready version were re-trained with the exact same fixed random seed of 42}. Consequently, the reported metrics may exhibit minor variations compared to preliminary runs, reflecting a more stable and reproducible convergence. \textbf{Importantly, we verified that these slight variations did not affect the results or conclusions of the subsequent experiments.}

\subsubsection{Results and Discussion}
Table \ref{tab:intrinsic_results} summarizes the performance of the five sequence-to-sequence models on the dialect normalization task using the test set. All models demonstrate good performance, but notable differences in effectiveness are observed.

\begin{table*}[t]
\centering
\begin{tabular}{lcccccc}
\hline
\textbf{Model} & \textbf{ROUGE-L} & \textbf{BLEU} & \textbf{METEOR} & \textbf{WER} & \textbf{CER} & \textbf{Parameters} \\
[0.5ex] 
\hline
BARTpho-word-base & 0.9167 & 0.7601 & 0.8625 & 0.1527 & 0.1049 & 150M \\
[0.5ex] 
ViT5-base & 0.9300 & 0.7934 & 0.8802 & 0.1340 & 0.0876 & 310M \\
[0.5ex]
BARTpho-syllable-base & 0.9218 & 0.7597 & 0.8627 & 0.1513 & 0.1045 & 132M \\
[0.5ex]
Vietnamese-correction-v2 & 0.9257 & 0.7723 & 0.8746 & 0.1416 & 0.0988 & 396M \\
[0.5ex]
mBART-large-50 & \textbf{0.9384} & \textbf{0.8166} & \textbf{0.8925} & \textbf{0.1226} & \textbf{0.0754} & 611M \\
[0.5ex]
\hline
\end{tabular}
\caption{Intrinsic Evaluation Results}
\label{tab:intrinsic_results}
\end{table*}

As shown in Table~\ref{tab:intrinsic_results}, \textbf{mBART-large-50} consistently outperforms all baselines across lexical and semantic metrics, confirming its strength for dialect normalization viewed as an intra-lingual translation task.

\textbf{ViT5-base} delivers comparable performance while using less than half the parameters, highlighting its parameter efficiency and suitability for real-world deployment.

\textbf{BARTpho} variants and \textbf{Vietnamese-correction-v2} yield moderate results, with the latter showing that pretraining on general correction tasks still transfers well. 

Overall, \textbf{mBART-large-50} provides the best performance, while \textbf{ViT5-base} offers a strong trade-off between accuracy and model size.

\section{Extrinsic Evaluation}
\label{sec:extrinsic_eval}
Beyond intrinsic metrics, extrinsic evaluations were conducted to demonstrate the practical utility of dialect normalization as a preprocessing step for real-world NLP tasks. These experiments assess whether normalization improves language processing at a semantic level, especially when dealing with non-standard Vietnamese inputs. 

\subsection{Machine Translation}
\label{subsec:machine_translation}
This section evaluates whether normalizing dialectal Vietnamese before translation improves English MT quality, under the hypothesis that standardization benefits systems trained on standard language.

\subsubsection{Experimental Design}
\label{subsec:mt_experimental_design}
Two parallel translation pipelines were designed for a dataset of 600 dialectal sentences, equally distributed across Northern, Central, and Southern regions (200 sentences per region), manually selected from ViDia2Std's test set. Each dialectal sentence had a human-translated English reference for evaluation.
\begin{itemize}
 \item \textbf{Direct Translation}: Dialectal sentences were translated directly into English using an MT system.
 \item \textbf{Translation after Normalization}: Dialectal sentences were first normalized to standard Vietnamese using the trained normalization model, then translated by the MT system.
\end{itemize}

\subsubsection{Evaluation Tool and Metric}
\label{subsec:machine_translation_metrics}
Traditional metrics (e.g., BLEU, ROUGE, METEOR) often fail to capture meaning preservation, especially in cases of paraphrasing or dialectal variation. We therefore adopt an \textit{LLM-as-a-Judge} approach, using Gemini 2.5 Flash to evaluate translations on two criteria: \textit{semantic completeness} and \textit{pragmatic accuracy}, producing a binary label: \texttt{ACCEPTED} or \texttt{UNACCEPTED}.

This approach is supported by recent work: \citet{sun-etal-2025-fine} find that BLEU underrepresents improvements in discourse coherence; \citet{kocmi-federmann-2023-large} show that GPT-based metrics (e.g., GEMBA) better align with human judgments than COMET; and \citet{fernandes2025llmsunderstandtranslationsevaluating} propose LLM-based QA to assess semantic retention directly.

These studies underscore a key limitation of traditional metrics—their poor handling of non-literal or creative translations \cite{li-etal-2025-generation,patil2025englishpleaseevaluatingmachine}. In contrast, LLMs offer contextual reasoning closer to human evaluators.

We report the \textbf{Acceptance Rate} as our primary metric:
\begin{equation}
\text{Acceptance Rate} = \frac{|\text{Accepted Samples}|}{|\text{Total Samples}|} \times 100\%
\end{equation}

\subsubsection{Results and Discussion}
As presented in Table \ref{tab:mt_overall_results_final_updated}, the machine translation evaluation results confirm that normalization improves translation quality across all systems. The most substantial gain was observed for Kimi-K2-Instruct (+12.84 percentage points), while even the top-performing system, Gemini 2.0 Flash, saw its acceptance rate increase from 61.83\% to 67.00\%. This demonstrates that normalization is a universally beneficial preprocessing step, reducing ambiguity for both commercial APIs and large language models.

\begin{table*}[t]
\centering
\begin{tabular}{lcccc}
\toprule
\textbf{Method} & \textbf{Total Samples} & \textbf{Accepted} & \textbf{Unaccepted} & \textbf{Acceptance Rate (\%)} \\
\midrule
Microsoft Azure AI Translator & 600 & 68 & 532 & 11.33 \\
Microsoft Azure AI Translator + Normalized & 600 & 134 & 466 & 22.33 \\
\midrule
Google Cloud Translation – Basic & 600 & 230 & 370 & 38.33 \\
Google Cloud Translation – Basic + Normalized & 600 & 275 & 325 & 45.83 \\
\midrule
DeepSeek-V3 & 600 & 327 & 273 & 54.50 \\
DeepSeek-V3 + Normalized & 600 & 379 & 221 & 63.17 \\
\midrule
Kimi-K2-Instruct & 600 & 308 & 292 & 51.33 \\
Kimi-K2-Instruct + Normalized & 600 & 385 & 215 & 64.17 \\
\midrule
Gemini 1.5 Flash & 600 & 282 & 318 & 47.00 \\
Gemini 1.5 Flash + Normalized & 600 & 342 & 258 & 57.00 \\
\midrule
Gemini 2.0 Flash & 600 & 371 & 229 & 61.83 \\
Gemini 2.0 Flash + Normalized & 600 & 402 & 198 & \textbf{67.00} \\
\bottomrule
\end{tabular}
\caption{Machine Translation Evaluation Results (Overall)}
\label{tab:mt_overall_results_final_updated}
\end{table*}

\subsubsection{Cross-System Consistency Analysis}
\label{subsec:cross-system_consistency}
To provide a more robust evaluation, this section analyzes the consistency of normalization\'s impact across all six translation systems. The trend in Table \ref{tab:consistency_analysis_final_6_models} is clear: the positive impact of normalization is more consistent than its negative effects. As the agreement threshold increases, the ratio of improved-to-worsened sentences grows significantly, from 1.70 for at least one system to 32:1 for at least four systems. Most notably, 3 sentences were unanimously improved by all six systems, while none were unanimously worsened. This provides strong evidence that the benefits are systematic and not system-specific artifacts, underscoring the reliable value of the normalization process.

\begin{table}[h]
\centering
\small
\begin{tabularx}{\linewidth}{>{\centering\arraybackslash}Xccc}
\toprule
\textbf{Threshold ($\geq$)} & \textbf{Improved} & \textbf{Worsened} & \textbf{Ratio} \\
\midrule
1 & 316 & 186 & 1.70 \\
2 & 152 & 48 & 3.17 \\
3 & 69 & 17 & 4.06 \\
4 & 32 & 1 & 32.00 \\
5 & 11 & 0 & $\infty$ \\
6 & 3 & 0 & $\infty$ \\
\bottomrule
\end{tabularx}
\caption{Consistency of Normalization Impact Across 6 Systems}
\label{tab:consistency_analysis_final_6_models}
\end{table}

\subsubsection{Per-System Impact Analysis}
\label{subsec:per-system_impact}
As detailed in Table \ref{tab:per_model_impact}, the improvement-to-worsening ratio varies across systems. Traditional services like Microsoft Azure AI Translator show a high ratio (4.00), indicating strong reliance on standard input. In contrast, the top-performing Gemini 2.0 Flash has the lowest ratio (1.66), suggesting it is more robust to dialectal variations but can still be adversely affected by suboptimal normalizations. DeepSeek-V3 shows a performance comparable to Google\'s translation service, with an improvement-to-worsening ratio of 2.04.

\begin{table}[h]
\small
\centering
\resizebox{\linewidth}{!}{
\begin{tabular}{lcc}
\toprule
\textbf{System} & \textbf{Improved} & \textbf{Worsened} \\
\midrule
Microsoft Azure AI Translator & 88 & 22 \\
Kimi-K2-Instruct & 125 & 48 \\
Gemini 1.5 Flash & 102 & 42 \\
Google Cloud Translation & 88 & 43 \\
DeepSeek-V3 & 102 & 50 \\
Gemini 2.0 Flash & 78 & 47 \\
\bottomrule
\end{tabular}
}
\caption{Per-System Sentence Improvement vs. Worsening}
\label{tab:per_model_impact}
\end{table}

\subsubsection{In-depth Analysis of Translation Regressions}
\label{subsec:translation_regressions}
While the aggregate results show a clear net benefit, a granular analysis of the regression cases—where a translation was downgraded from \texttt{ACCEPTED} to \texttt{UNACCEPTED} post-normalization—is essential. We manually analyzed all 252 instances of regression across the six translation systems to categorize the root cause of the failure. The results, summarized in Table \ref{tab:regression_analysis_summary}, reveal that the regressions are not primarily caused by the normalization process itself, but by limitations in the downstream components of the pipeline.

\begin{table}[h]
\centering
\small
\renewcommand{\arraystretch}{1.2}
\begin{tabularx}{\linewidth}{l c X}
\toprule
\textbf{Error Category}
 & \textbf{Count (\%)}
 & \textbf{Description} \\
\midrule
MT Model Fragility
 & 118 (46.8\%)
 & The MT system fails to translate a correctly normalized sentence. \\
LLM Evaluator Noise
 & 90 (35.7\%)
 & The evaluator provides inconsistent ratings for translations of equivalent quality. \\
Normalization Model Error
 & 44 (17.5\%)
 & The normalization model incorrectly alters the meaning of the source sentence. \\
\midrule
\textbf{Total}
 & \textbf{252 (100\%)}
 & \\
\bottomrule
\end{tabularx}
\caption{Categorization of Translation Regression Causes Across All Systems (percentages shown in parentheses next to counts)}
\label{tab:regression_analysis_summary}
\end{table}

Our analysis identifies three primary sources of error, with their new ranking:

\begin{enumerate}
 \item \textbf{MT Model Fragility (46.8\%):} The largest source of regression involves failures of the downstream MT system. Even when provided with a perfectly standard Vietnamese sentence, the MT system sometimes produced a translation that was significantly worse than the one generated from the original dialect. These failures suggest that MT systems may be overfitted to certain linguistic patterns and lack the robustness to handle valid, albeit less common, phrasings.

 \item \textbf{LLM Evaluator Noise (35.7\%):} The second-largest category is the inherent stochasticity and phrasal sensitivity of the LLM-as-a-Judge. In numerous cases, the direct and normalized translations were semantically identical, yet received different verdicts. This highlights a key challenge in automated evaluation: the judge\'s preference for a specific phrasing can be misinterpreted as a difference in quality.

 \item \textbf{Normalization Model Error (17.5\%):} The normalization model itself was the least frequent source of regression. These errors typically occurred when the model failed to preserve nuanced semantics.
\end{enumerate}

Crucially, the fact that over 82\% of regressions are attributable to MT model fragility and evaluator noise provides strong evidence for the reliability of our normalization approach. These regressions are not systematic flaws but rather isolated artifacts of the current evaluation and translation pipeline. This reinforces the conclusion from our consistency analysis (Table \ref{tab:consistency_analysis_final_6_models}): the benefits of normalization are systematic, while the regressions are largely system-specific and non-systematic.

\subsection{Sentiment Analysis}
\label{subsec:sentiment_analysis}
To assess the impact of dialect normalization on sentiment analysis performance, a pre-trained Vietnamese sentiment analysis model was evaluated before and after applying dialect normalization. This design aimed to quantify the improvement in the model's ability to correctly understand the emotional tone of dialectal utterances.

\subsubsection{Experimental Design}
\label{subsubsec:sentiment_analyis_experimental_design}
A test set of dialectal sentences was used, which were automatically labeled for sentiment (Positive, Negative, Neutral) using Gemini 2.0 Flash and deepseek-ai/DeepSeek-V3. The automatic labeling achieved a Cohen’s kappa of 0.7082, which is considered acceptable and indicative of substantial agreement on a 300-sample consensus set. A pre-trained sentiment analysis model (5CD-AI/Vietnamese-Sentiment-visobert \cite{Visobert2024}) was then applied to classify sentiment under two scenarios: (1) direct analysis on original dialectal input, and (2) analysis on input normalized by the model.

\subsubsection{Evaluation Metrics}
\label{subsubsec:sentiment_analysis_metrics}
Standard classification metrics were used: Accuracy, Precision, Recall, and F1-score.

\subsubsection{Results and Discussion}
\label{subsubsec:sentiment_analysis_results}
Table~\ref{tab:sa_comparison} demonstrates that dialect normalization substantially improves sentiment analysis performance. Accuracy increased from 50.59\% to 62.13\%, and macro F1 rose from 0.48 to 0.58.

Performance improved across all sentiment classes. F1-score for NEGATIVE rose from 0.59 to 0.72, NEUTRAL from 0.37 to 0.45, and POSITIVE from 0.47 to 0.58. These gains indicate that dialect normalization helps disambiguate regional lexical forms that often confuse standard models, especially for nuanced sentiment expressions.

Normalization corrected a significant number of prediction errors, with the majority in the NEGATIVE and NEUTRAL categories. This confirms that dialectal ambiguity most affects less polarized sentiments.

\begin{table}[t]
\centering
\resizebox{\linewidth}{!}{%
\begin{tabular}{lccc|ccc}
\toprule
& \multicolumn{3}{c|}{\textbf{Before}} & \multicolumn{3}{c}{\textbf{After}} \\
\cmidrule(r){2-4} \cmidrule(l){5-7}
\textbf{Class} & P & R & F1 & P & R & F1 \\
\midrule
NEGATIVE & 0.89 & 0.45 & 0.59 & \textbf{0.86↓} & \textbf{0.62↑} & \textbf{0.72↑} \\
NEUTRAL & 0.47 & 0.30 & 0.37 & \textbf{0.48↑} & \textbf{0.42↑} & \textbf{0.45↑} \\
POSITIVE & 0.32 & 0.87 & 0.47 & \textbf{0.44↑} & \textbf{0.82↓} & \textbf{0.58↑} \\
\midrule
\textbf{Accuracy} & \multicolumn{3}{c|}{0.51} & \multicolumn{3}{c}{\textbf{0.62↑}} \\
\textbf{Macro Avg} & 0.56 & 0.54 & 0.48 & \textbf{0.59↑} & \textbf{0.62↑} & \textbf{0.58↑} \\
\textbf{Weighted Avg} & 0.68 & 0.51 & 0.52 & \textbf{0.69↑} & \textbf{0.62↑} & \textbf{0.63↑} \\
\bottomrule
\end{tabular}
}
\caption{Sentiment analysis before vs.\ after dialect normalization. Bold↑ indicates improvement; ↓ indicates decline.}
\label{tab:sa_comparison}
\end{table}

\subsubsection{Analysis of Changes}
\label{subsubsec:sentiment_analysis_analysis_of_changes}
To further understand the impact of dialect normalization, we analyzed how the sentiment model's predictions changed for each sentence. A total of 1603 sentences were evaluated. The analysis reveals that the normalization process had a net positive effect, correcting more errors than it introduced.

As summarized in Table~\ref{tab:analysis_summary}, out of the 1603 sentences, a significant number of prediction errors were corrected.
Specifically:
\begin{itemize}
    \item \textbf{Improvement (Corrected Errors)}: 265 sentences (16.53\% of the dataset) saw their incorrect predictions become correct after normalization. This indicates that the model was better able to understand the intended sentiment of the dialectal text.
    \item \textbf{Regression (New Errors)}: Only 80 sentences (4.99\% of the dataset) were misclassified after normalization, having been correctly classified before.
\end{itemize}
This results in a net improvement of 185 sentences, leading to an overall accuracy increase of 11.54\% ($0.6213 - 0.5059$).

Analyzing these changes by sentiment class provides deeper insights. The normalization process was highly effective for the \textbf{NEGATIVE} class, with a net improvement of 160 sentences. The \textbf{NEUTRAL} class also saw a positive net effect of 41 sentences. However, for the \textbf{POSITIVE} class, the normalization had a slightly negative net effect, introducing 28 new errors while only correcting 12. This suggests that the dialectal forms for positive sentiment may be more complex or ambiguous, and further refinement is needed for this specific class.

The overall success rate of the normalization, calculated as the ratio of corrected errors to total changed predictions, is high at 76.81\% ($\frac{265}{265+80}$). This strong performance confirms that dialect normalization is a highly effective pre-processing step for improving sentiment analysis of Vietnamese dialectal text.

\begin{table}[h]
    \centering
    \footnotesize 
    \begin{tabular}{|l|c|c|}
        \hline
        \textbf{Type of Change} & \textbf{Count} & \textbf{Percentage (\%)} \\
        \hline
        Incorrect $\to$ Correct & 265 & 16.53\% \\
        Correct $\to$ Incorrect & 80 & 4.99\% \\
        Correct $\to$ Correct & 731 & 45.60\% \\
        Incorrect $\to$ Incorrect & 527 & 32.88\% \\
        \hline
        \textbf{Total} & \textbf{1603} & \textbf{100.00\%} \\
        \hline
    \end{tabular}
    \caption{Comparison: Improvement vs. Regression after Normalization}
    \label{tab:analysis_summary}
\end{table}

\section{Limitations}
\label{sec:limitations}
While the ViDia2Std corpus and our sequence-to-sequence models establish robust benchmarks for Vietnamese dialect normalization, the work also reveals several inherent challenges. A primary challenge is the tendency toward \textbf{over-normalization}, where models sometimes "over-normalize" expressions, resulting in the loss of critical pragmatic or stylistic cues embedded in the original dialect. This failure manifests as the inaccurate substitution of formal terms or specialized terminology with common standard equivalents. This issue is confirmed through our in-depth analysis of machine translation regressions (Table~\ref{tab:regression_analysis_summary}), where \textit{Normalization Model Error} (failure to preserve nuanced semantics) was the root cause in 17.5\% of the analyzed regression instances.

A second limitation, as noted by reviewers, is the reliance on a \textbf{single LLM-as-a-Judge} for the extrinsic evaluation. This approach, while scalable, carries an inherent risk of evaluator bias and stochasticity. This concern is quantitatively supported by our own manual analysis in Table~\ref{tab:regression_analysis_summary}. We found that \textit{LLM Evaluator Noise} was the second-largest cause of regression, accounting for a significant 35.7\% of all cases where a translation was downgraded. This finding confirms that the LLM judge's stochasticity is a valid concern. Future work should incorporate a parallel human evaluation study to establish a clearer correlation and quantify the precise agreement between the LLM judge and human assessments.

\section{Conclusion and Future Work}
\label{sec:conclusion_and_future Work}
We introduced ViDia2Std, the first large-scale, manually annotated Vietnamese dialect-to-standard corpus spanning all 63 provinces. With over 13,000 sentence pairs from authentic social media, it offers a new benchmark for dialectal diversity. Experiments show that sequence-to-sequence models—especially mBART-large-50—can effectively normalize dialectal input, leading to significant improvements in downstream tasks like machine translation and sentiment analysis. These findings highlight the value of dialect-aware preprocessing in Vietnamese NLP.

Future work includes (1) expanding the corpus to cover more sources (e.g., spoken transcripts, forums) and genres, and (2) addressing over-normalization, where pragmatic or stylistic cues are lost. We aim to develop context-sensitive models that can distinguish between dialectal terms needing normalization and colloquialisms that should be preserved. This ensures clarity without erasing linguistic nuance.

\section*{Ethics Statement}
\label{sec:ethics_statement}
The data for the ViDia2Std corpus was collected exclusively from public Facebook news fanpages. No private user data was accessed. To protect user privacy, all collected data was anonymized during the preprocessing pipeline. All metadata, such as usernames, user IDs, and mentions, was stripped from the comments. The dataset is intended solely for non-commercial linguistic research. This work was completed before Vietnam’s June 12, 2025 reorganization (Resolution No. 202/2025/QH15), which reduced provincial‑level units from 63 to 34 (effective July 1, 2025); all maps, indexing and analysis use the 63‑province structure current at project design.

\section*{Acknowledgments}
\label{sec:acknowledgments}
This research was supported by The VNUHCM-University of Information Technology’s Scientific Research Support Fund. We would like to thank the anonymous reviewers, the Senior Program Committee (SPC), and the Area Chair (AC) of AAAI-26. Their insightful comments and constructive feedback significantly improved the quality of this paper.

\bibliography{aaai2026}
\end{document}